\documentclass[10pt,twocolumn,letterpaper]{article}

\usepackage{cvpr}      
\usepackage{makecell}
\usepackage[utf8]{inputenc}
\usepackage{geometry}

\definecolor{cvprblue}{rgb}{0.21,0.49,0.74}
\usepackage[pagebackref,breaklinks,colorlinks,allcolors=cvprblue]{hyperref}

\def\paperID{3967} 
\def\confName{CVPR}
\def\confYear{2026}

\title{GA-Drive: Geometry-Appearance Decoupled Modeling for Free-viewpoint Driving Scene Generation}

\author{
    Hao Zhang$^{1*}$, \quad Lue Fan$^{1,2*}$, \quad Qitai Wang$^{2}$, \quad Wenbo Li$^{3}$, \\
    Zehuan Wu$^{4}$, \quad Lewei Lu$^{4}$, \quad Zhaoxiang Zhang$^{2\dagger}$, \quad Hongsheng Li$^{1\dagger}$ \\[0.5em]
    {\small $^{1}$MMLab, CUHK \qquad $^{2}$CASIA} \\
    {\small $^{3}$Shanghai Jiaotong University \qquad $^{4}$SenseTime Research}
}

\date{} 

\begin{document}

\twocolumn[{%
    \renewcommand\twocolumn[1][]{#1}%
    \maketitle
    \vspace{-0.3in}
    \begin{center}
      \includegraphics[width=0.9\linewidth]{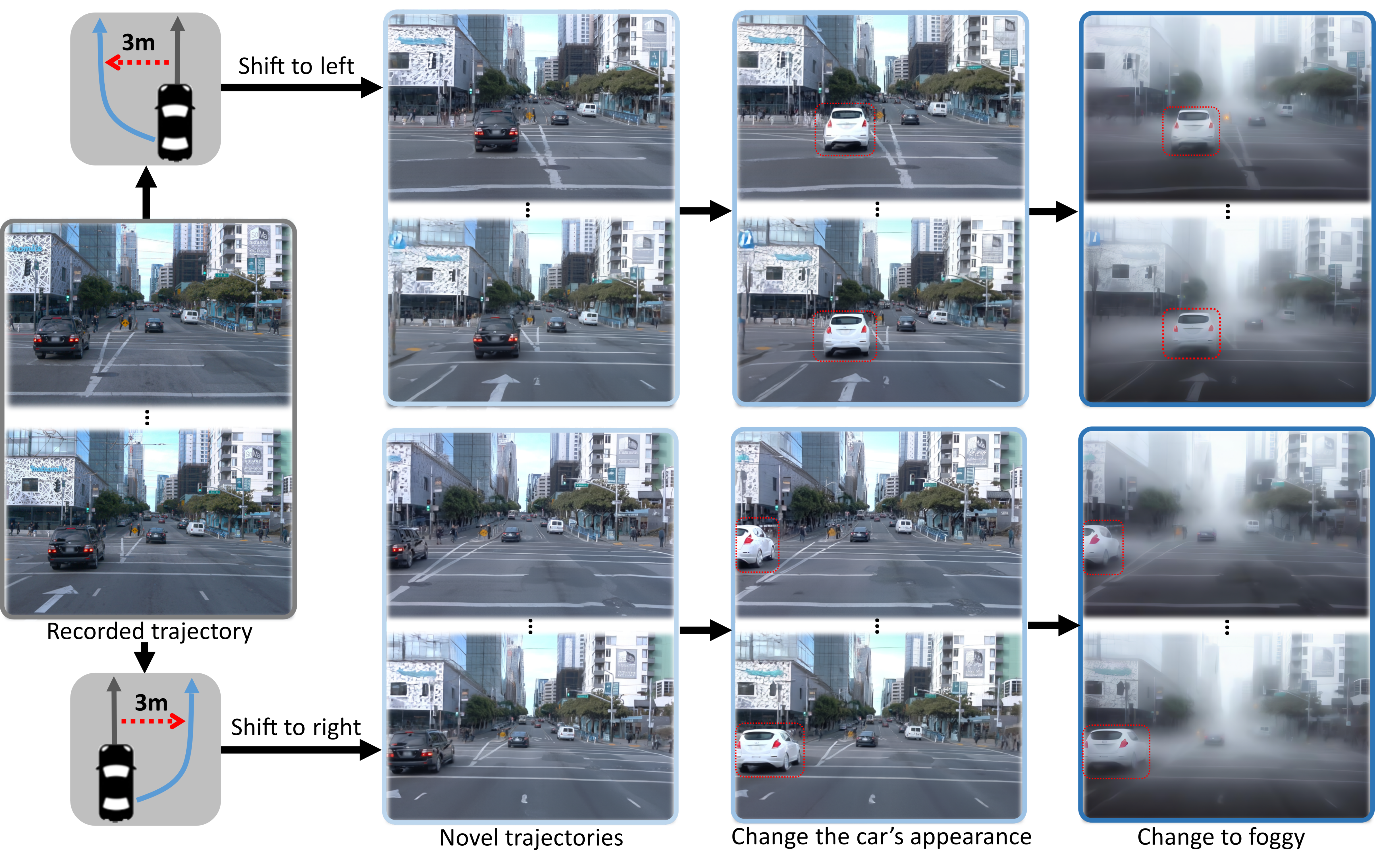}
      \vspace{-0.1in}
      \captionsetup{hypcap=false} 
      \captionof{figure}{GA-Drive generates high-quality and novel views along shifted trajectories. By decoupling geometry from appearance, our method allows flexible editing of appearance, such as altering the car’s appearance and changing the weather to fog, without impacting the underlying geometry. The \textcolor{red}{red} bounding boxes show that appearance editing is consistent across different trajectories.}
      \label{fig:teaser}
    \end{center}
}]

\begin{abstract}
A free-viewpoint, editable, and high-fidelity driving simulator is crucial for training and evaluating end-to-end autonomous driving systems. In this paper, we present GA-Drive, a novel simulation framework capable of generating camera views along user-specified novel trajectories through Geometry-Appearance Decoupling and Diffusion-Based Generation. Given a set of images captured along a recorded trajectory and the corresponding scene geometry, GA-Drive synthesizes novel pseudo-views using geometry information. These pseudo-views are then transformed into photorealistic views using a trained video diffusion model.
In this way, we decouple the geometry and appearance of scenes.
An advantage of such decoupling is its support for appearance editing via state-of-the-art video-to-video editing techniques, while preserving the underlying geometry, enabling consistent edits across both original and novel trajectories. 
Extensive experiments demonstrate that GA-Drive substantially outperforms existing methods in terms of NTA-IoU, NTL-IoU, and FID scores. 

\end{abstract}    
\section{Introduction}

Developing a photorealistic driving simulator with scene editing capabilities is critical for training and evaluating end-to-end autonomous driving systems. 
Reconstruction-based simulators, such as PVG~\cite{chen2023periodic}, OmniRe~\cite{chen2024omnire}, and S3Gaussian~\cite{huang2024s3gaussian} reconstruct driving scenes from recorded trajectories. 
However, they can only render high-quality views along pre-recorded trajectories, limiting their ability to support free-viewpoint simulation.
FreeSim~\cite{fan2024freesim}, DriveDreamer4D~\cite{zhao2024drive}, and ReconDreamer~\cite{Ni2024ReconDreamerCW} use diffusion-based generative models to synthesize novel views and reconstruct 4D dynamic scenes from both real and synthesized images.
However, these methods often struggle to maintain view consistency or preserve details when combining generated and real views, resulting in suboptimal reconstructions and blurred visual results. Additionally, they generally lack the capability for scene editing.
Recent world models, such as DriveDreamer-2~\cite{zhao2025drivedreamer} and MaskGWM~\cite{ni2025maskgwm}, attempt to model dynamic scenes by generating videos. 
However, unlike reconstruction-based methods, they can only control the camera pose in a coarse manner, and 3D consistency cannot be guaranteed. As a result, they are unable to support accurate free-viewpoint simulation.

To this end, we propose GA‑Drive, a novel photorealistic driving simulator that supports free‑viewpoint simulation and appearance editing via geometry–appearance decoupling and diffusion‑based generation.
Our key insight is to decouple geometry from appearance. In driving scenes with limited viewpoints, RGB images cannot uniquely determine 3D geometry. Fitting appearance sacrifices geometric accuracy by introducing floaters or distortions, which cause obvious artifacts on novel trajectories. By separating these components, we enable the geometry module to focus on accurate 3D structure while delegating appearance synthesis to a diffusion model.


Our geometry module builds on OmniRe~\cite{chen2024omnire}, a 4D reconstruction method. We introduce dense depth supervision and depth-distortion regularization to obtain more accurate geometry. To provide the diffusion model with sufficient geometric and appearance cues for generating 3D-consistent novel views, we leverage this accurate reconstruction to synthesize novel pseudo-views that encode correct 3D spatial relationships and occlusion patterns. Specifically, we cast rays from the target novel pose into the reconstructed 3D space, forming a point cloud, then project these points onto captured source views to retrieve their colors. Although the resulting pseudo-views may contain holes or occluded regions, they serve as a strong prior that guides our video diffusion model to generate photorealistic, 3D-consistent novel views.

However, training our video diffusion model presents a key challenge. Ideally, with multiple trajectories in the same scene, we could use views from trajectory A to synthesize pseudo-views of trajectory B, then supervise the model with real views from trajectory B. Unfortunately, public driving datasets offer only a single trajectory per scene. While captured views can serve as ground-truth supervision, there is no second trajectory to provide sources for synthesizing pseudo-view conditions.

To address this challenge, we introduce a pseudo-view simulation pipeline that simulates training pairs from the single available trajectory, circumventing the need for multiple simultaneously recorded trajectories. Simulated pseudo-views serve as conditioning frames for the video diffusion model, while the original captured views serve as ground-truth supervision. Moreover, this approach enables scalable training on large-scale datasets such as Waymo~\cite{Sun_2020_CVPR} and OpenDV~\cite{gao2024vista}, equipping the model to handle diverse driving scenarios, including challenging conditions like fog and snow.


Since all geometric information is encapsulated in the 4D geometry representation while recorded views serve solely as appearance sources, geometry and appearance are effectively decoupled.
After decoupling, our method produces geometry-consistent, photorealistic appearances for novel trajectory views. This decoupling also enables flexible appearance editing without affecting the underlying geometry. Once the appearance of the original recorded videos is modified, the appearance of all generated novel views across arbitrary trajectories is updated accordingly, ensuring global consistency. The editing of the original videos can be achieved using state-of-the-art video editing methods, such as VACE~\cite{vace}, Instruct-V2V~\cite{cheng2023consistent}, or other advanced techniques.

We summarize our contributions as follows
\begin{enumerate}
    \item \textbf{Geometry-Appearance Decoupling}: A novel framework that separates geometry and appearance for diffusion-based generation in driving simulators. This feature enables flexible appearance editing.
    \item \textbf{Pseudo-view Simulation Pipeline}: A pipeline that generates training pairs for video diffusion models without requiring datasets with multiple simultaneously recorded trajectories.
    \item The proposed framework substantially outperforms existing methods for the novel trajectory synthesis task in terms of NTA-IoU, NTL-IoU, and FID scores.

\end{enumerate}
\section{Related Works}

\textbf{Reconstruction-Based Driving Simulators.}  
Recent advancements in 3D scene reconstruction have been driven by Neural Radiance Fields (NeRF)~\cite{mildenhall2020nerf} and 3D Gaussian Splatting (3DGS)~\cite{kerbl20233d}. NeRF models scenes as continuous volumetric radiance fields, while 3DGS represents scenes using anisotropic Gaussian clouds. Building upon these foundations, numerous works have extended NeRF and 3DGS to unbounded, dynamic driving environments~\cite{chen2023periodic,huang2024s3gaussian,wu2023mars,guo2023streetsurf,turki2023suds,yang2023unisim,tonderski2024neurad,zhou2024drivinggaussian,yan2024street}. For example, OmniRe~\cite{chen2025omnire} introduces non-rigid modeling for dynamic actors such as pedestrians and cyclists. However, these reconstruction-based approaches are inherently limited by the original camera trajectory. Since they lack observations beyond the recorded viewpoints, they struggle to render high-quality novel views from significantly off-trajectory positions.

\textbf{Generation-Reconstruction Hybrid Driving Simulators.}  
To overcome the limitations imposed by the original viewpoint distribution, recent approaches have explored hybrid models that combine generative techniques with scene reconstruction for novel view synthesis (NVS). FreeSim~\cite{fan2024freesim} progressively incorporates generated views into reconstruction, beginning near the recorded trajectory and expanding outward. Similarly, DriveDreamer4D~\cite{zhao2024drive} and ReconDreamer~\cite{Ni2024ReconDreamerCW} use generative outputs as supervision signals to improve reconstruction quality. While these methods aim to blend the strengths of generative and reconstruction-based models, they still face challenges. Generative models often struggle to preserve consistent 3D structure across viewpoints, leading to under-converged geometry and blurred scene details. Moreover, as these methods ultimately rely on 3DGS as the scene representation, they inherit the same limitations in editability and flexibility as pure reconstruction-based approaches.

\textbf{Diffusion-Based Driving Simulators.}  
Diffusion-based generative models have also been explored for driving simulation. For instance, Vista~\cite{gao2024vista} and MaskGWM~\cite{ni2025maskgwm} generate driving videos from text prompts. However, these models do not construct an explicit 4D representation of the scene, limiting the ability to directly control camera poses. FreeVS~\cite{wang2024freevs} addresses this by incorporating LiDAR point clouds as priors to guide video generation, allowing explicit control of the target camera pose via the LiDAR projection matrix. Nevertheless, FreeVS cannot recover 3D content not captured by the LiDAR sensor, such as distant or elevated structures, which constrains the completeness of the generated scenes.

\begin{figure*}
    \centering
    \includegraphics[width=\linewidth]{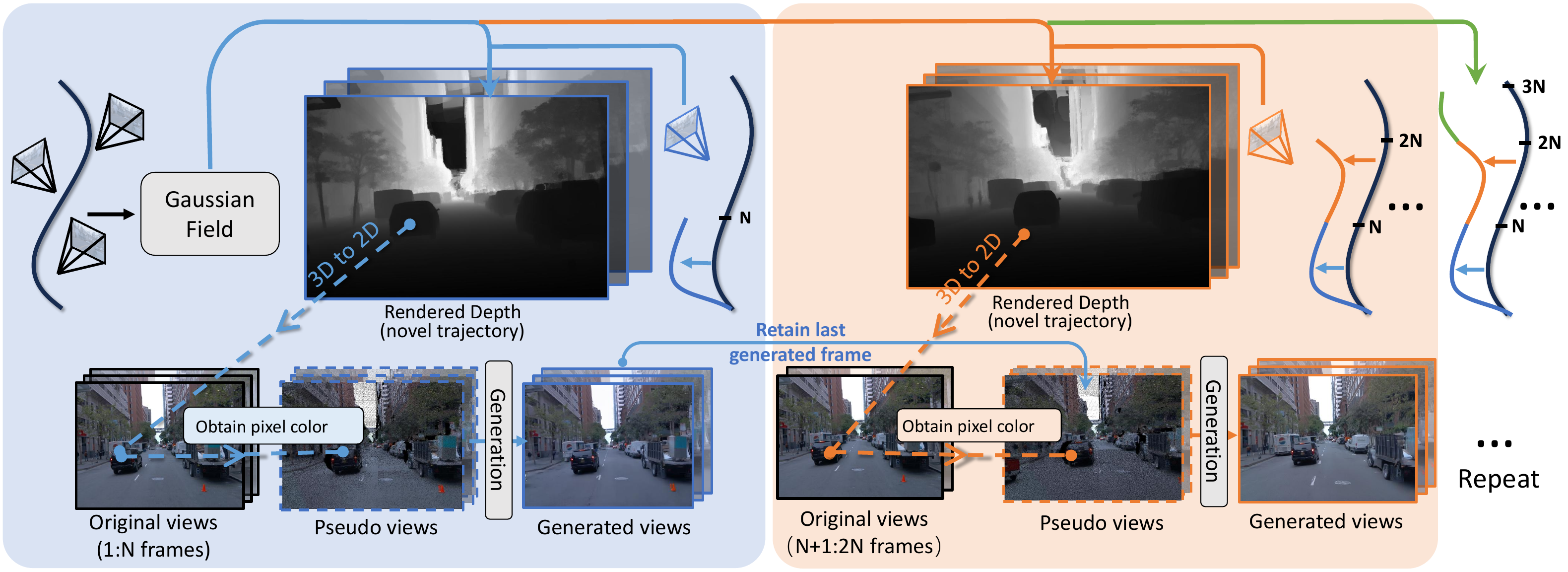}
    \vspace{-0.25in}
    \caption{{\bf The overview framework of GA-Drive.} Pseudo views are created by casting rays from novel poses into 3D space with rendered depth maps to generate a 3D point cloud, which is then projected onto recorded views to sample color information. These pseudo views are subsequently transformed into photorealistic frames using our trained video diffusion model (referred to as Generation). The black curved line indicates the recorded camera trajectory. By iteratively applying the video diffusion model to each segment, our method can generate photorealistic novel views of unlimited length. }
    \label{fig:framework}
    \vspace{-0.2in}
\end{figure*}

\section{Method}
Fig.\ref{fig:framework} shows the overview of our framework. The geometry module is presented in~\cref{sec:geometry}. We then describe the Geometry-Guided Pseudo-View Synthesis in~\cref{sec:Pseudo} and the Video Diffusion Model in~\cref{sec:Diffusion}. Finally, the Pseudo-View Simulation Pipeline for training is detailed in~\cref{sec:Training}.

\subsection{4D Geometry Reconstruction}\label{sec:geometry}

We adopt OmniRe~\cite{chen2025omnire}, a Gaussian scene graph representation, as our 4D reconstruction framework. In real‑world captures, limited viewpoints and sensor imperfections mean colors alone cannot uniquely determine 3D geometry.
This ambiguity often causes Gaussian-splatting reconstructions to exhibit geometric distortions, such as floating Gaussians, when fitting to observed colors.
Since we decouple the geometry and appearances, our interest lies solely in the geometric structure of the driving scene.
Thus, we apply the following two techniques to improve geometric accuracy.

\noindent \textbf{Dense depth supervision} 
The original LiDAR-based depth supervision in OmniRe is inherently sparse and limited to the bottom half of the view due to sensor constraints. To address this limitation, we introduce a dense depth supervision loss $\mathcal{L}_{d\_esti}$ that provides geometric guidance in regions lacking LiDAR observations. This dense supervision is particularly critical in scenarios with limited viewpoint variation, where greater geometric ambiguity makes reliable 3D geometry reconstruction challenging. Our approach leverages Marigold~\cite{ke2025marigold} to estimate relative depth maps for each input image. We then align these estimates to absolute scale by projecting LiDAR point clouds onto the image plane, obtaining sparse yet accurate depth values at the projected pixels. A linear transformation (scale and shift) is estimated to minimize the mean squared error between the relative depth and absolute LiDAR depth at overlapping locations. The loss $\mathcal{L}_{d\_esti}$ measures the L1 distance between the aligned depth estimates and the rendered depth from gaussian field. This dense supervision provides geometric constraints across the entire image, significantly improving reconstruction quality where direct LiDAR coverage is unavailable. Further implementation details are provided in the supplementary materials.


\noindent \textbf{Depth distortion regularization} 
We further incorporate the depth distortion loss proposed by Mip-nerf 360~\cite{barron2022mip}, which encourages the weight distribution along each ray to be more concentrated by minimizing the distances between ray-splat intersections. With this loss, the Gaussians are encouraged to cluster more closely around the scene surface, avoiding the formation of floating Gaussians.
Please refer to the supplementary materials for an explanation of how we implement depth distortion regularization into the OmniRe backbone without modifying the original OmniRe CUDA code.




\subsection{Geometry-Guided Pseudo-View Synthesis} \label{sec:Pseudo}
Given a 4D reconstruction of the driving scene, we cast a ray from each pixel into 3D space based on the pixel's depth, which is derived from the 4D reconstruction. 
At a particular timestamp, assume we have access to $N$ recorded images 
, along with their corresponding camera poses and depth maps rendered from the Gaussian field. Given a target camera pose along a novel trajectory, our objective is to synthesize a pseudo-view by leveraging these $N$ source images.

\noindent\textbf{Problem Setup} \quad Given $N$ source views $\{( \mathbf{I}_i, \mathbf{K}_i, \mathbf{T}_i,  \mathbf{D}_i)\}_{i=1}^N$, where $I_i$ is the known RGB image, $\mathbf{K}_i \in \mathbf{R}^{3 \times 3}$ is the camera intrinsic matrix, $\mathbf{T}_i \in SE(3)$ is the camera-to-world transformation, $\mathbf{D}_i \in \mathbf{R}^{H \times W}$ is the depth map rendered from the gaussian field, and a target camera defined by $(\mathbf{K}_t, \mathbf{T}_t, \mathbf{D}_t)$. Our goal is to synthesize a pseudo-view $I_t^{\text{pseudo}}$ that approximates the appearance of the scene from the target viewpoint.
 

\paragraph{Step 1: Unproject target view depth to 3D world coordinates.}

For each pixel $(u, v)$ in the target image with depth $D_t(u, v)$, we compute the corresponding 3D point $\mathbf{x}_w$ in world space

\vspace{-0.25in}
\begin{equation}
\mathbf{x}_w = \mathbf{T}_t \begin{bmatrix} \mathbf{D}_t(u, v)\, \mathbf{K}_t^{-1} \mathbf{p}_{uv} \\ 1 \end{bmatrix}, \quad \mathbf{p}_{uv} = \begin{bmatrix} u \\ v \\ 1 \end{bmatrix}
\end{equation}


Repeat for all pixels to obtain a set of 3D points $\mathcal{X} = \{\mathbf{x}_w^j\}_{j=1}^{H \cdot W}$. We denote the color of a 3D point $\mathbf{x}_w$ as $c_{\mathbf{x}_w}$. Initially, each point is colored black; we will assign them specific colors later. 

\paragraph{Step 2: Project 3D points into source views.}

Each point $\mathbf{x}_w$ is projected to each source view $i$:
\begin{equation}
\label{eq:projection}
\begin{split}
    \mathbf{x}_c^{(i)} &= \mathbf{T}_i^{-1} \begin{bmatrix} \mathbf{x}_w \\ 1 \end{bmatrix}, \\
    \mathbf{p}^{(i)} = \mathbf{K}_i \cdot \mathbf{x}_c^{(i)}[:3], &\quad \mathbf{u}^{(i)} = \left( \frac{p_x^{(i)}}{p_z^{(i)}}, \frac{p_y^{(i)}}{p_z^{(i)}} \right).
\end{split}
\end{equation}

Only points with $p_z^{(i)} > 0$ that fall within the height (H) and width (W) bounds of source view $i$ are considered.

\vspace{-0.1in}
\paragraph{Step 3: Sample colors from recorded views and check visibility.}
We determine the visibility of a point and sample its corresponding color using the following equation. 

\begin{figure}
    \centering
    \includegraphics[width=\linewidth]{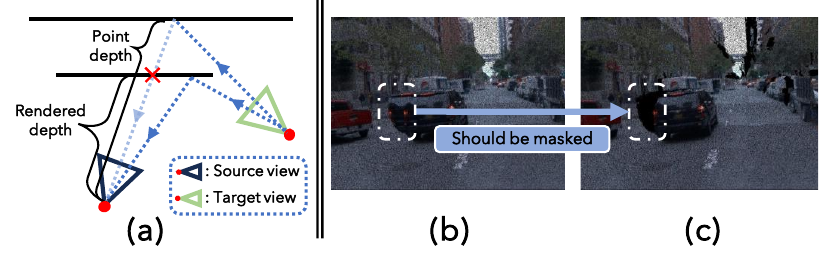}
    \vspace{-0.25in}
    \caption{(a) illustrates the 2D version of the visibility check. Since the rendered depth $<$ the point depth, the point is occluded by a surface and should be masked. (b) shows an incorrect pseudo view without applying the visibility check, while (c) presents the correct result with visibility properly handled. }
    \vspace{-0.2in}
    \label{fig:visibility_check}
\end{figure}

\vspace{-0.2in}
\begin{equation}
    c_{\mathbf{x}_w} = \begin{cases}I_i(\mathbf{u}^{(i)}) &  \mbox{if}  \left| \mathbf{D}_i(\mathbf{u}^{(i)}) - \mathbf{x}_c^{(i)}[2] \right| < \delta. \\
    \mathbf{0} & \mbox{otherwise} \end{cases}
    \label{eq:sample_color}
\end{equation}

where $\mathbf{D}_i(\mathbf{u}^{(i)})$ is the depth rendered from the gaussian field. $\mathbf{x}_c^{(i)}[2]$ is the depth of the 3D points in the camera coordinate. If  $\mathbf{x}_c^{(i)}[2] > \mathbf{D}_i(\mathbf{u}^{(i)})$, the point lies behind the surface and is considered occluded. In this case, the point should not be assigned a color and remains black. We introduce a tolerance parameter $\delta$ to account for potential inaccuracies in the geometry. A 2D illustration of this occlusion handling is shown in Fig.~\ref{fig:visibility_check} for clarity. Fig.~\ref{fig:visibility_check}(b) displays a pseudo-view without visibility checks, while Fig.~\ref{fig:visibility_check}(c) includes visibility check.

\paragraph{Step 4: Color Aggregation.} 
We execute Equation~\ref{eq:sample_color} for all recorded views in descending order of their distance to the target camera. For each 3D point $\mathbf{x}_w \in \mathcal{X}$, if it is visible in multiple views,  we assign its color from the nearest valid source view. 
After assigning colors to all 3D points in $\mathcal{X}$, we project each point back to the target view to form the pseudo view $I_t^{\text{pseudo}}$. 








\subsection{Segment-wise Video Diffusion Model} \label{sec:Diffusion}
\begin{figure}
    \centering
    \includegraphics[width=\linewidth]{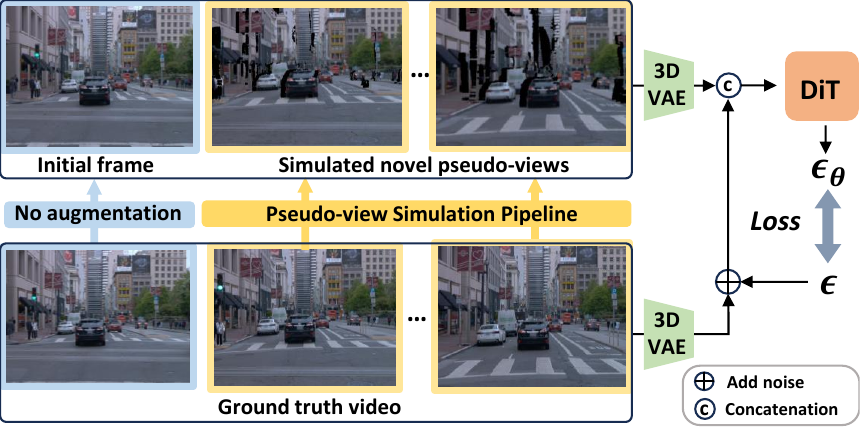}
    \vspace{-0.25in}
    \caption{{\bf The architecture of our video diffusion model.} Our segment-wise video diffusion model generates videos conditioned on the novel pseudo-views.}
    \label{fig:diffusion_model}
    \vspace{-0.2in}
\end{figure}

We generate photorealistic novel views from novel pseudo-views that may contain missing or occluded regions by leveraging a video diffusion model. To generate long video trajectories, we employ a segment-wise approach where each video segment is conditioned on a sequence of frames: the first conditioning frame is set to the last generated frame from the previous segment to ensure temporal consistency, while the remaining conditioning frames are the corresponding pseudo-views along the trajectory. The overall architecture is illustrated in Fig.\ref{fig:diffusion_model}.

Our video diffusion model builds upon the I2V architecture from CogVideoX~\cite{yang2024cogvideox}.
To incorporate conditioning frames in the video diffusion model, we encode them using the same 3D variational autoencoder (VAE) as CogVideoX. The resulting latent features are then concatenated with Gaussian noise along the channel dimension. 
A segment of conditioning frames is represented as a tensor of shape $T \times H \times W \times 3$, where $T$ is the number of frames, $H$ and $W$ are the height and width, and $3$ denotes RGB channels. The 3D VAE encodes this to a latent tensor of size $\frac{T}{q} \times \frac{H}{p} \times \frac{W}{p} \times 16$, with $q$ and $p$ as temporal and spatial downsampling factors and $16$ latent channels. We sample a gaussian noise tensor of the same shape, concatenate it with the encoded features along the channel dimension, yielding a combined tensor of shape $\frac{T}{q} \times \frac{H}{p} \times \frac{W}{p} \times 32$, which is input to the diffusion model’s denoising network. Model fine-tuning follows the DDPM framework~\cite{ho2020denoising}, as in CogVideoX. In our setup, $T=16$, $H=640$, $W=960$, and the text prompt is fixed as ``A realistic autonomous driving scene.''


The generation of novel trajectory views begins from a camera pose with a known image (recorded or generated). For each segment, the first conditioning frame is this known image, which ensures temporal consistency, while the remaining conditioning frames are the corresponding pseudo-views that provide geometric and appearance guidance for the target viewpoints. Once a segment is completed, its final generated frame becomes the first conditioning frame for the next segment, and the process repeats.
During training, the first conditioning frame is the ground-truth frame from the recorded trajectory, while the remaining conditioning frames are simulated pseudo-views synthesized via the pipeline in~\cref{sec:Training}. Please see the supplementary materials for more details.

\begin{figure}
    \centering
    \includegraphics[width=\linewidth]{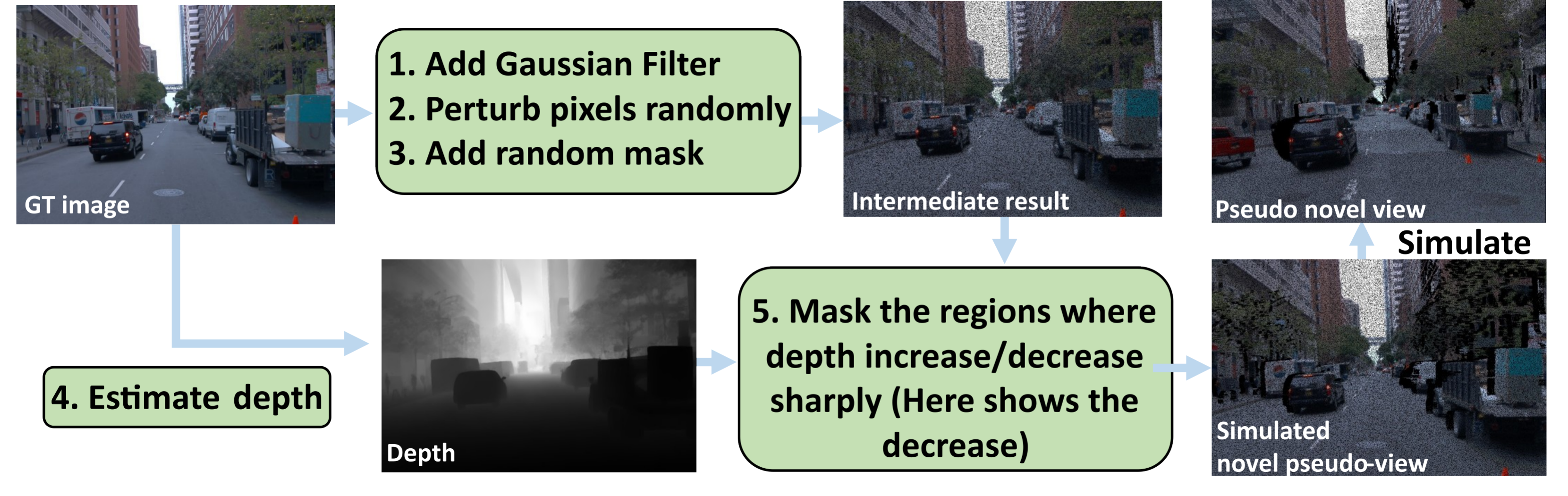}
    \vspace{-0.25in}
    \caption{{\bf The pseudo-view simulation pipeline.} We design this pseudo-view simulation pipeline to simulate the characteristic patterns of the novel pseudo-views.}
    \label{fig:pipeline}
    \vspace{-0.1in}
\end{figure}

\subsection{Pseudo-view Simulation Pipeline} \label{sec:Training}

We propose a pseudo-view simulation pipeline that simulates the characteristic patterns observed in novel pseudo-views, enabling us to train our segment-wise video diffusion model directly on single-trajectory datasets. This eliminates the need for multi-trajectory recordings, which are unavailable in public driving datasets since a vehicle cannot traverse multiple paths simultaneously.

\paragraph{The characteristic patterns of the novel pseudo-views. } As shown in Fig.~\ref{fig:pipeline}, the novel pseudo-views exhibit several characteristic patterns: (1) Random masks, which arise due to numerical inaccuracies during the unprojection and reprojection steps. Because floating-point operations are not perfectly precise, some pixels may not be projected back to their original locations, resulting in empty (black) pixels; (2) Local blending artifacts, where inaccurate depth estimation may cause a pixel to be sampled from a nearby, but incorrect, location in the source view; and (3) Occulded regions, typically caused by occlusions or blocked areas detected by visibility check. 

\paragraph{The pipeline to simulate characteristic patterns. }  Our pseudo-view simulation pipeline shown in Fig.~\ref{fig:pipeline} includes the following steps: (1) A Gaussian filter is applied to reduce high-frequency noise and simulate the blurring introduced by projection artifacts; (2) For each pixel, we randomly blend it with a nearby pixel to reproduce the local blending behavior described in pattern (2); (3) We randomly mask some pixels to simulate the random masks observed in pattern (1); and (4) to simulate pattern (3), we mask the regions where the estimated depth increase/decrease sharply. The estimated depth is obtained via a monocular depth estimation model, Marigold~\cite{ke2025marigold}. This simulation pipeline enables the video diffusion model to be trained on single-trajectory datasets.

Extensive experiments demonstrate that there is no noticeable gap between novel pseudo-views and simulated novel pseudo-views. We train the diffusion model on the Waymo training dataset~\cite{Sun_2020_CVPR} along with 700 videos featuring diverse weather conditions selected from OpenDV~\cite{gao2024vista}—a large-scale dataset of driving videos collected from YouTube. Our efficient simulation pipeline, which processes each frame in just three seconds on an A100 GPU, makes such large-scale training feasible.

\section{Experiment}

\subsection{Appearance Editing}

Since our method disentangles geometry from appearance, appearance editing can be performed easily by modifying the recorded images. We can leverage SOTA video editing methods, such as VACE~\cite{vace}, Instruct-V2V~\cite{cheng2023consistent}, or other techniques to edit the appearance of recorded images. 
Once the recorded images are altered, all novel trajectories generated from them automatically reflect these changes, as the pseudo-views sample colors directly from the edited images.
This ensures consistent appearance editing across arbitrary camera trajectories.
For example, we can modify the weather in a scene simply by providing a text prompt using Instruct-V2V. Additionally, the mv2v function in the VACE model allows us to alter the appearance of cars. Thanks to recent advances in powerful video editing methods, our approach can leverage these tools to enable versatile editing of driving scenes.
Please refer to Fig.~\ref{fig:teaser} for qualitative results demonstrating appearance edits. We also encourage you to check the supplementary for additional results.

\begin{figure*}
    \centering

    \includegraphics[width=\linewidth]{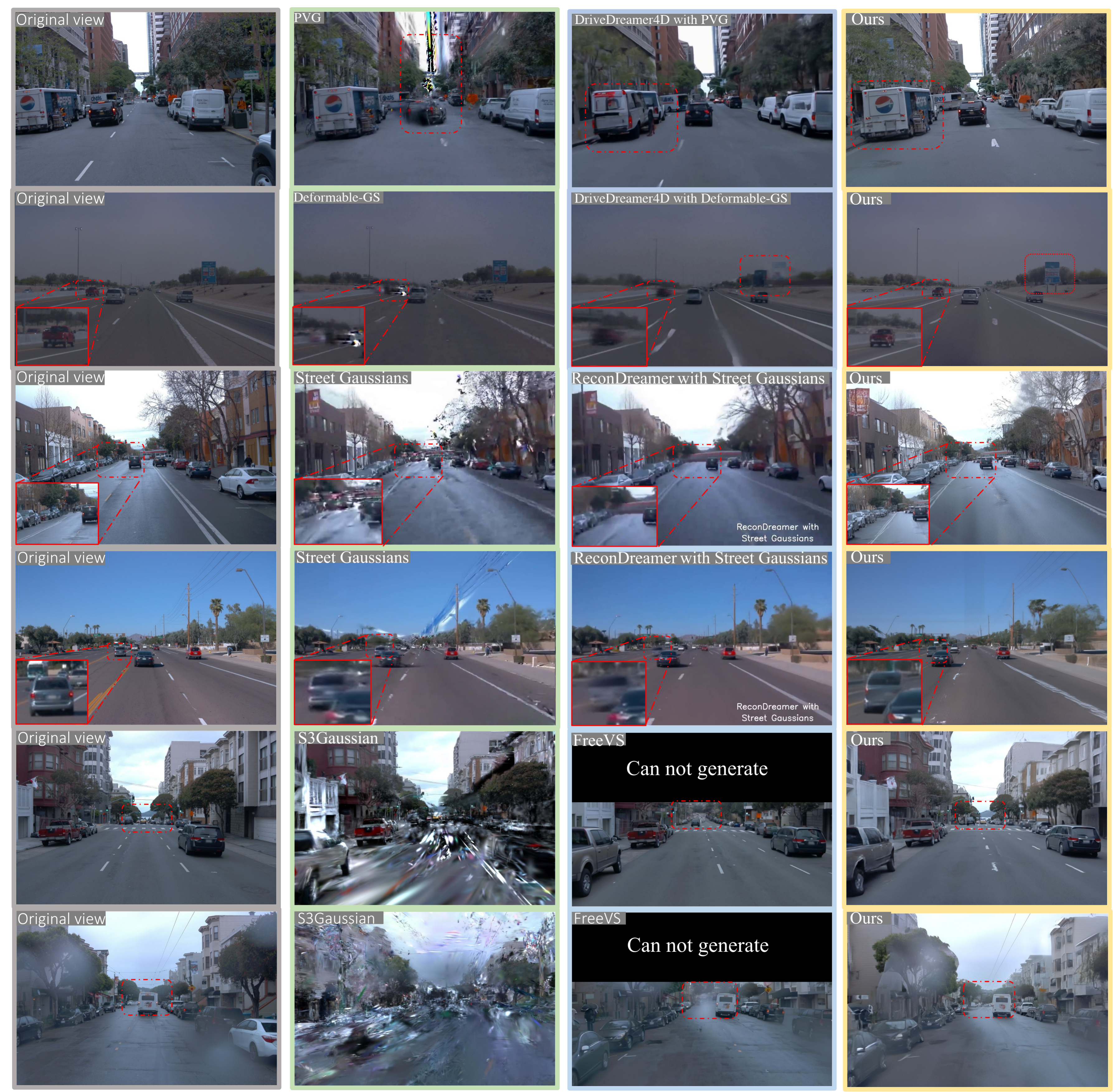}

    \caption{Qualitative comparison of novel trajectories with a 3-meter lane shift. In the comparison, the fourth row is shifted to the right, while the others are shifted to the left.}

    \label{fig:comparison}
\end{figure*}

\subsection{Comparison}
\label{sec:comparison}

We conduct all experiments using the same test dataset as ReconDreamer~\cite{Ni2024ReconDreamerCW}, selected from the Waymo validation set. Specific scene IDs and frame numbers are provided in the supplementary materials. 

\begin{table*}
\centering
\resizebox{\textwidth}{!}{
\begin{tabular}{l|ccc|ccc|ccc|c}
\toprule
Models &  PVG & \makecell{DD4D w/.\\ PVG} & \makecell{Recon. w/. \\  PVG} & $S^3$Gauss. & \makecell{DD4D w/.\\  $S^3$Gauss.} & \makecell{Recon. w/. \\  $S^3$Gauss.} & Deform.-GS & \makecell{DD4D w/.\\ Deform.-GS} & \makecell{Recon. w/. \\  Deform.-GS} & Ours \\
\midrule
\midrule
NTA-IoU$\uparrow$ & 0.256 & 0.438  & 0.464 & 0.175 & 0.495 & 0.413 & 0.240 & 0.335 & 0.443 & {\bf 0.558} \\
NTL-IoU$\uparrow$  & 50.70 & 53.06 & 53.21 & 49.05 & 53.42 & 51.62 & 51.62 & 52.93 & 53.78 & {\bf 56.83}  \\
FID$\downarrow$  & 105.29 & 71.52 & 74.32 & 124.90 & 66.93 & 123.61 & 92.24 & 77.32 & 76.24 & {\bf 49.77}\\

\bottomrule
\end{tabular}}
\caption{Quantitative comparison in a lane-change scenario where the trajectory gradually shifts 4 m to the left.}

\label{tab:table_comparison}
\end{table*}

\paragraph{Quantitative comparison} To evaluate the ability of our model to generate high-fidelity novel trajectories, we quantitatively compare it against several state-of-the-art baselines: PVG~\cite{chen2023periodic}, S3Gaussian~\cite{huang2024s3gaussian}, DriveDreamer4D~\cite{zhao2024drive}, ReconDreamer~\cite{Ni2024ReconDreamerCW}, and Deformable-GS~\cite{yang2023deformable3dgs}. Both DriveDreamer4D and ReconDreamer have three implementations, using PVG, S3Gaussian, and Deformable-GS as their underlying scene representations. We report results for each variant independently. Since FreeVS~\cite{wang2024freevs} only generates partial novel views, we include it in qualitative comparisons only. We adopt the metrics Novel Trajectory Agent IoU (NTA-IoU) and Novel Trajectory Lane IoU (NTL-IoU), proposed in DriveDreamer4D, to evaluate spatial consistency of vehicles and lane boundaries in novel views. Additionally, we compute the Fréchet Inception Distance (FID) between original and novel trajectory images to assess photorealism. To better simulate realistic driving scenarios, we evaluate methods on a lane change trajectory with a lateral shift of 0.1 meters per frame over 40 frames, producing a total lateral shift  of 4 m. Quantitative results are shown in ~\Cref{tab:table_comparison}. Our method achieves the highest scores on both NTA-IoU and NTL-IoU, indicating superior ability to synthesize recognizable and accurately positioned vehicles and lane boundaries. Furthermore, our method yields the lowest FID, demonstrating its strength in generating photorealistic images compared to other baselines.

\paragraph{Qualitative comparison} As DriveDreamer4D and ReconDreamer have not released their models, we perform qualitative comparisons using the examples provided on their official websites. We replicate their experimental settings for a fair comparison. All experiments are based on a 3-meter leftward shift, except for the second ReconDreamer comparison, which is evaluated on a 3-meter rightward shift. As illustrated in Fig.~\ref{fig:comparison}, our method produces novel trajectories with greater consistency to the original views compared to DriveDreamer4D, especially in regions highlighted by red boxes. This improvement arises because DriveDreamer4D conditions only on the first frame, leading to shape inconsistencies in subsequent frames. Compared to ReconDreamer, our model achieves higher fidelity and temporal coherence in the generated views. ReconDreamer adopts a generation-reconstruction hybrid framework, which can introduce blurring due to temporal inconsistencies in generated details. Unlike FreeVS, which is limited to regions visible to the LiDAR sensor, our method generates full novel views. Lastly, the four reconstruction methods (PVG, Deformable-GS, S3Gaussian, and Street Gaussians ~\cite{yan2024street}) exhibit obvious artifacts on novel trajectories compared to other approaches.


\subsection{Ablation study}
\begin{figure}
    \centering
    \includegraphics[width=\linewidth]{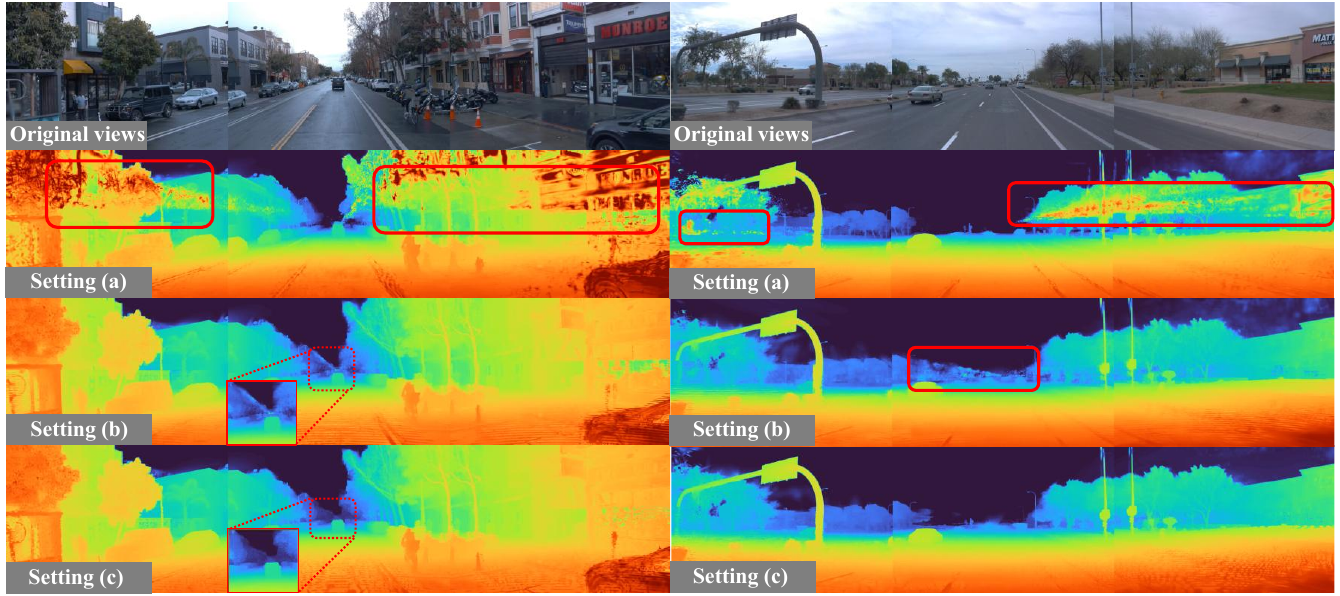}
    \vspace{-0.25in}
    \caption{We conduct an ablation study on Dense Depth Supervision and Depth Distortion Regularization. Specifically, setting (a) refers to training OmniRe with the original configuration, (b) adds Dense Depth Supervision, and (c) includes both Dense Depth Supervision and Depth Distortion Regularization. Setting (c) produces the cleanest geometry.}
    \label{fig:Ablation_aligned}
\end{figure}

\paragraph{Ablation study on 4D Geometry Reconstruction}

We introduce Dense Depth Supervision and Depth Distortion Regularization in ~\cref{sec:geometry} to enhance geometry reconstruction quality. To evaluate their effectiveness, we train OmniRe models on the Waymo dataset under the following configurations:

\begin{enumerate}[label=(\alph*)]
\item Training OmniRe with the original setting;
\item Training OmniRe with Dense Depth Supervision;
\item Training OmniRe with both Dense Depth Supervision and Depth Distortion Regularization.
\end{enumerate}

As illustrated in Fig.~\ref{fig:Ablation_aligned}, the OmniRe model trained with the original setting (a) exhibits substantial depth estimation errors, especially in regions lacking LiDAR data—such as the upper half of the scene or distant background areas. Applying Dense Depth Supervision (b) significantly improves geometry in these regions, though some Gaussian floaters remain visible. Incorporating both Dense Depth Supervision and Depth Distortion Regularization (c) yields the highest quality geometry, particularly in LiDAR-sparse areas, and further reduces the presence of Gaussian floaters. These results demonstrate the effectiveness of our proposed methods in improving geometry reconstruction.



\paragraph{Ablation study on Segment-wise Video Diffusion Model}
\begin{figure}
    \centering
    \includegraphics[width=\linewidth]{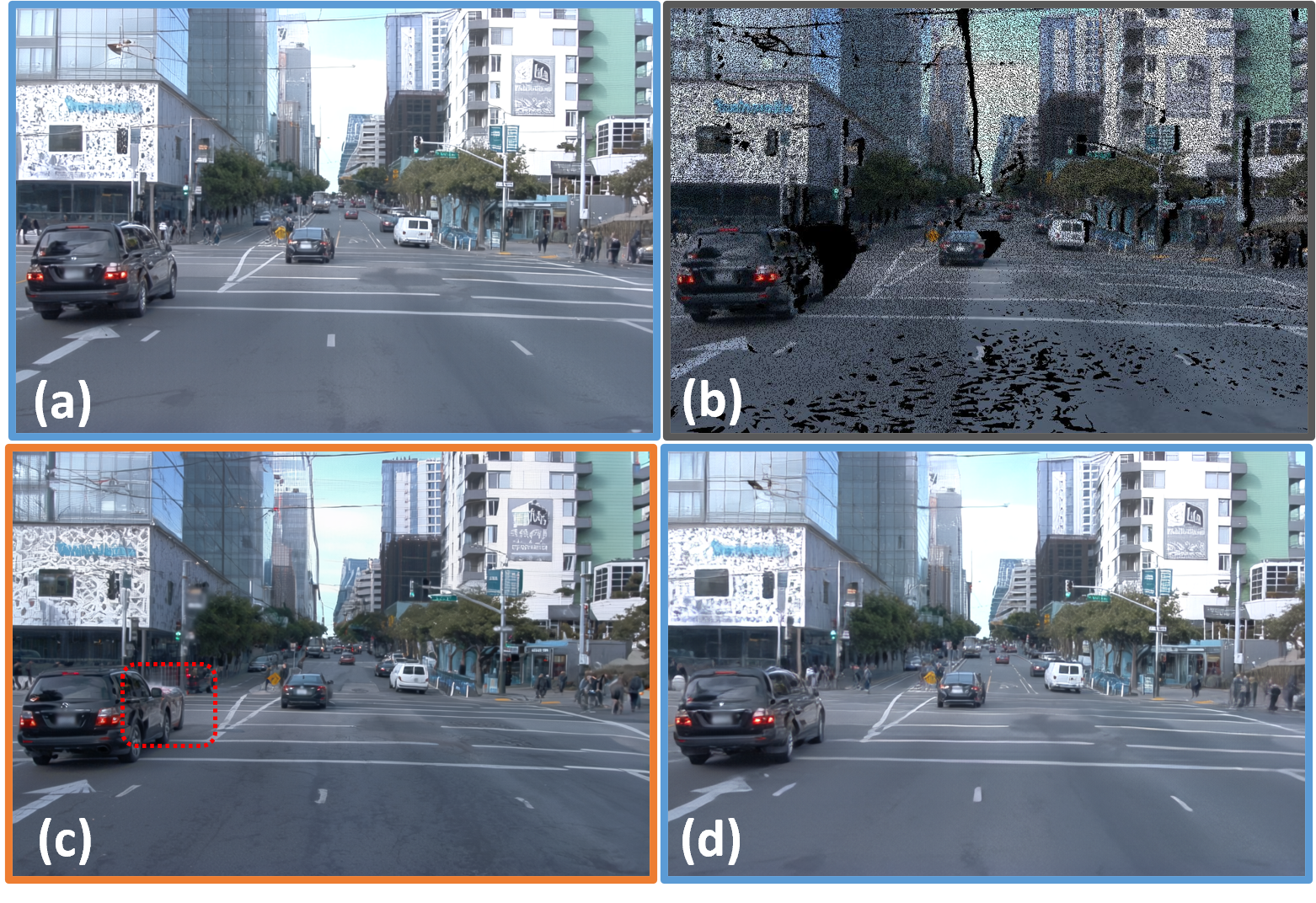}
    \vspace{-0.25in}
    \caption{Ablation study on segment-wise video generation: (a) last frame of the previous segment; (b) first pseudo-novel view of the next segment; (c) view generated by the ablation model; (d) view generated by our default model. Compared to (c), (d) provides a much more coherent transition with (a).}
    \label{fig:Ablation_seg}
\end{figure}

To assess the importance of segment-wise conditioning for coherent transitions between segments, we train a variant where the first conditioning frame of each segment is also a simulated pseudo-view rather than the last generated frame from the previous segment.
We test both models on two consecutive segments. Fig.~\ref{fig:Ablation_seg} shows: (a) the last frame of the previous segment, (b) the first novel pseudo-view of the next segment, (c) the view generated from (b) by the ablation model, and (d) the view generated from (b) by our default model. Our method yields smoother transitions and better visual consistency across segments, especially in regions with missing content, demonstrating the effectiveness of segment-wise anchoring.

\begin{table}[t]
  \centering
  \small
  \setlength{\tabcolsep}{6pt} 
  \begin{tabular}{lccc}
    \toprule
    Corruption & NTA-IoU $\uparrow$ & NTL-IoU $\uparrow$ & FID $\downarrow$ \\
    \midrule
    --                 & 0.558 & 56.83 & 49.77 \\
    $-20\%$ Gaussians  & 0.555 & 56.54 & 50.67 \\
    $-40\%$ Gaussians  & 0.542 & 56.02 & 51.75 \\
    $-80\%$ Gaussians  & 0.535 & 54.79 & 50.86 \\
    $\pm 0.1$ m noise  & 0.545 & 56.62 & 52.13 \\
    $\pm 0.2$ m noise  & 0.538 & 56.24 & 50.97 \\
    \bottomrule
  \end{tabular}
  \vspace{-0.1in}
  \caption{Performance under different corruptions. The first row is our default model without corruption}
  \label{tab:corruption_table}
  \vspace{-0.2in}
\end{table}

\paragraph{Robustness of Segment-wise Video Diffusion Model}
We evaluate the robustness of our Segment‑wise video diffusion model by corrupting the underlying 3D Gaussian representation to synthesize pseudo‑novel views with severe artifacts, and then requiring our video diffusion model to restore them. The corruption comprises two schemes: (i) randomly dropping a subset of Gaussian primitives and (ii) adding i.i.d. translation noise to each Gaussian’s x, y, and z coordinates. On the same test set as ~\cref{sec:comparison}, ~\Cref{tab:corruption_table} shows that removing up to 80\% of Gaussians or injecting uniformly sampled translation noise within ±0.2 m yields only marginal changes in final metrics.
These results demonstrate strong robustness, which we attribute to the model’s generalizability and the effectiveness of our pseudo‑view simulation pipeline. Please check the supplementary for additional visual results.






\subsection{Limitations and Conclusion}


Our method relies on 4D geometry reconstruction from OmniRe. While the Gaussian reconstruction backbone may introduce minor geometric inaccuracies because of rolling shutter distortion, particularly during high-speed driving, our diffusion model demonstrates strong robustness to such imperfections and still produces high-quality results. Additionally, our current framework focuses on appearance editing rather than geometry editing. Future work could explore integrating rolling shutter correction into the 3DGS rasterizer and extending the framework to support geometry editing capabilities.

We present GA-Drive, a novel framework for photorealistic driving simulation that enables novel trajectory synthesis and appearance editing. By decoupling appearance from geometry, GA-Drive overcomes key limitations of previous simulators, such as blurring, limited editability, and inconsistency with original views. Our improved 4D reconstruction yields more accurate geometry. A dedicated pseudo-view simulation pipeline enables training a video diffusion model without costly 4D reconstruction, allowing photorealistic novel view generation across diverse driving scenarios. To enable long-term novel view synthesis, we introduce a segment-wise video generation framework that ensures coherence across segments. Collectively, these innovations establish GA-Drive as a powerful tool for training and evaluating autonomous driving systems in diverse, photorealistic, and interactive environments.

{
    \small
    \bibliographystyle{ieeenat_fullname}
    \bibliography{main}

@String(CVPR= {IEEE Conf. Comput. Vis. Pattern Recog.})

@String(ECCV= {Eur. Conf. Comput. Vis.})

@String(AAAI = {AAAI})

@String(CVPR  = {CVPR})

@String(ECCV  = {ECCV})

@inproceedings{
  chen2025omnire,
  title={OmniRe: Omni Urban Scene Reconstruction},
  author={Ziyu Chen and Jiawei Yang and Jiahui Huang and Riccardo de Lutio and Janick Martinez Esturo and Boris Ivanovic and Or Litany and Zan Gojcic and Sanja Fidler and Marco Pavone and Li Song and Yue Wang},
  booktitle={The Thirteenth International Conference on Learning Representations},
  year={2025}
}

@article{fan2024freesim,
  title={FreeSim: Toward Free-viewpoint Camera Simulation in Driving Scenes},
  author={Fan, Lue and Zhang, Hao and Wang, Qitai and Li, Hongsheng and Zhang, Zhaoxiang},
  journal={arXiv preprint arXiv:2412.03566},
  year={2024}
}

@article{chen2023periodic,
  title={Periodic Vibration Gaussian: Dynamic Urban Scene Reconstruction and Real-time Rendering},
  author={Chen, Yurui and Gu, Chun and Jiang, Junzhe and Zhu, Xiatian and Zhang, Li},
  journal={arXiv:2311.18561},
  year={2023},
}

@inproceedings{yan2024street,
    title={Street Gaussians: Modeling Dynamic Urban Scenes with Gaussian Splatting}, 
    author={Yunzhi Yan and Haotong Lin and Chenxu Zhou and Weijie Wang and Haiyang Sun and Kun Zhan and Xianpeng Lang and Xiaowei Zhou and Sida Peng},
    booktitle={ECCV},
    year={2024}
}

@article{yang2024cogvideox,
  title={CogVideoX: Text-to-Video Diffusion Models with An Expert Transformer},
  author={Yang, Zhuoyi and Teng, Jiayan and Zheng, Wendi and Ding, Ming and Huang, Shiyu and Xu, Jiazheng and Yang, Yuanming and Hong, Wenyi and Zhang, Xiaohan and Feng, Guanyu and others},
  journal={arXiv preprint arXiv:2408.06072},
  year={2024}
}

@article{ho2020denoising,
  title={Denoising diffusion probabilistic models},
  author={Ho, Jonathan and Jain, Ajay and Abbeel, Pieter},
  journal={Advances in neural information processing systems},
  volume={33},
  pages={6840--6851},
  year={2020}
}

@inproceedings{Ni2024ReconDreamerCW,
    title={ReconDreamer: Crafting World Models for Driving Scene Reconstruction via Online Restoration},
    author={Chaojun Ni and Guosheng Zhao and Xiaofeng Wang and Zheng Zhu and Wenkang Qin and Guan Huang and Chen Liu and Yuyin Chen and Yida Wang and Xueyang Zhang and Yifei Zhan and Kun Zhan and Peng Jia and Xianpeng Lang and Xingang Wang and Wenjun Mei},
    booktitle = {arXiv}, 
    year={2024},
    url={https://arxiv.org/abs/2411.19548}
}

@inproceedings{mildenhall2020nerf,
  title     = {NeRF: Representing Scenes as Neural Radiance Fields for View Synthesis},
  author    = {Ben Mildenhall and Pratul P. Srinivasan and Matthew Tancik and Jonathan T. Barron and Ravi Ramamoorthi and Ren Ng},
  year      = {2020},
  booktitle = {ECCV}
}

@article{kerbl20233d,
  title={3D Gaussian Splatting for Real-Time Radiance Field Rendering},
  author={Kerbl, Bernhard and Kopanas, Georgios and Leimk{\"u}hler, Thomas and Drettakis, George},
  journal={ACM Transactions on Graphics},
  volume={42},
  number={4},
  year={2023}
}

@inproceedings{wu2023mars,
  title={Mars: An instance-aware, modular and realistic simulator for autonomous driving},
  author={Wu, Zirui and Liu, Tianyu and Luo, Liyi and Zhong, Zhide and Chen, Jianteng and Xiao, Hongmin and Hou, Chao and Lou, Haozhe and Chen, Yuantao and Yang, Runyi and others},
  booktitle={CAAI International Conference on Artificial Intelligence},
  pages={3--15},
  year={2023},
  organization={Springer}
}

@article{guo2023streetsurf,
  title={StreetSurf: Extending Multi-view Implicit Surface Reconstruction to Street Views},
  author={Guo, Jianfei and Deng, Nianchen and Li, Xinyang and Bai, Yeqi and Shi, Botian and Wang, Chiyu and Ding, Chenjing and Wang, Dongliang and Li, Yikang},
  journal={arXiv preprint arXiv:2306.04988},
  year={2023}
}

@inproceedings{yang2023unisim,
  title={UniSim: A Neural Closed-Loop Sensor Simulator},
  author={Yang, Ze and Chen, Yun and Wang, Jingkang and Manivasagam, Sivabalan and Ma, Wei-Chiu and Yang, Anqi Joyce and Urtasun, Raquel},
  booktitle={Proceedings of the IEEE/CVF Conference on Computer Vision and Pattern Recognition},
  pages={1389--1399},
  year={2023}
}

@inproceedings{turki2023suds,
  title={SUDS: Scalable Urban Dynamic Scenes},
  author={Turki, Haithem and Zhang, Jason Y and Ferroni, Francesco and Ramanan, Deva},
  booktitle={Proceedings of the IEEE/CVF Conference on Computer Vision and Pattern Recognition},
  pages={12375--12385},
  year={2023}
}

@inproceedings{tonderski2024neurad,
  title={Neurad: Neural rendering for autonomous driving},
  author={Tonderski, Adam and Lindstr{\"o}m, Carl and Hess, Georg and Ljungbergh, William and Svensson, Lennart and Petersson, Christoffer},
  booktitle={Proceedings of the IEEE/CVF Conference on Computer Vision and Pattern Recognition},
  pages={14895--14904},
  year={2024}
}

@inproceedings{zhou2024drivinggaussian,
  title={Drivinggaussian: Composite gaussian splatting for surrounding dynamic autonomous driving scenes},
  author={Zhou, Xiaoyu and Lin, Zhiwei and Shan, Xiaojun and Wang, Yongtao and Sun, Deqing and Yang, Ming-Hsuan},
  booktitle={Proceedings of the IEEE/CVF Conference on Computer Vision and Pattern Recognition},
  pages={21634--21643},
  year={2024}
}

@article{huang2024s3gaussian,
        title={S3Gaussian: Self-Supervised Street Gaussians for Autonomous Driving},
        author={Huang, Nan and Wei, Xiaobao and Zheng, Wenzhao and An, Pengju and Lu, Ming and Zhan, Wei and Tomizuka,    Masayoshi and Keutzer, Kurt and Zhang, Shanghang},
        journal={arXiv preprint arXiv:2405.20323},
        year={2024}
      }

@inproceedings{zhao2024drive,
    title={DriveDreamer4D: World Models Are Effective Data Machines for 4D Driving Scene Representation}, 
    author={Guosheng Zhao and Chaojun Ni and Xiaofeng Wang and Zheng Zhu and Xueyang Zhang and Yida Wang and Guan Huang and Xinze Chen and Boyuan Wang and Youyi Zhang and Wenjun Mei and Xingang Wang},
    booktitle = {arXiv}, 
    journal={arxiv arXiv preprint arXiv:2410.13571},
    year={2024},
}

@article{yang2023deformable3dgs,
    title={Deformable 3D Gaussians for High-Fidelity Monocular Dynamic Scene Reconstruction},
    author={Yang, Ziyi and Gao, Xinyu and Zhou, Wen and Jiao, Shaohui and Zhang, Yuqing and Jin, Xiaogang},
    journal={arXiv preprint arXiv:2309.13101},
    year={2023}
}

@article{wang2024freevs,
  title={Freevs: Generative view synthesis on free driving trajectory},
  author={Wang, Qitai and Fan, Lue and Wang, Yuqi and Chen, Yuntao and Zhang, Zhaoxiang},
  journal={arXiv preprint arXiv:2410.18079},
  year={2024}
}

@article{chen2024omnire,
    title={OmniRe: Omni Urban Scene Reconstruction},
    author={Chen, Ziyu and Yang, Jiawei and Huang, Jiahui and Lutio, Riccardo de and Esturo, Janick Martinez and Ivanovic, Boris and Litany, Or and Gojcic, Zan and Fidler, Sanja and Pavone, Marco and Song, Li and Wang, Yue},
    journal={arXiv preprint arXiv:2408.16760},
    year={2024}
}

@article{ni2025maskgwm,
  title={Maskgwm: A generalizable driving world model with video mask reconstruction},
  author={Ni, Jingcheng and Guo, Yuxin and Liu, Yichen and Chen, Rui and Lu, Lewei and Wu, Zehuan},
  journal={arXiv preprint arXiv:2502.11663},
  year={2025}
}

@InProceedings{Sun_2020_CVPR, author = {Sun, Pei and Kretzschmar, Henrik and Dotiwalla, Xerxes and Chouard, Aurelien and Patnaik, Vijaysai and Tsui, Paul and Guo, James and Zhou, Yin and Chai, Yuning and Caine, Benjamin and Vasudevan, Vijay and Han, Wei and Ngiam, Jiquan and Zhao, Hang and Timofeev, Aleksei and Ettinger, Scott and Krivokon, Maxim and Gao, Amy and Joshi, Aditya and Zhang, Yu and Shlens, Jonathon and Chen, Zhifeng and Anguelov, Dragomir}, title = {Scalability in Perception for Autonomous Driving: Waymo Open Dataset}, booktitle = {Proceedings of the IEEE/CVF Conference on Computer Vision and Pattern Recognition (CVPR)}, month = {June}, year = {2020} }

@misc{ke2025marigold,
  title={Marigold: Affordable Adaptation of Diffusion-Based Image Generators for Image Analysis},
  author={Bingxin Ke and Kevin Qu and Tianfu Wang and Nando Metzger and Shengyu Huang and Bo Li and Anton Obukhov and Konrad Schindler},
  year={2025},
  eprint={2505.09358},
  archivePrefix={arXiv},
  primaryClass={cs.CV}
}

@inproceedings{gao2024vista,
 title={Vista: A Generalizable Driving World Model with High Fidelity and Versatile Controllability}, 
 author={Shenyuan Gao and Jiazhi Yang and Li Chen and Kashyap Chitta and Yihang Qiu and Andreas Geiger and Jun Zhang and Hongyang Li},
 booktitle={Advances in Neural Information Processing Systems (NeurIPS)},
 year={2024}
}

@article{vace,
    title = {VACE: All-in-One Video Creation and Editing},
    author = {Jiang, Zeyinzi and Han, Zhen and Mao, Chaojie and Zhang, Jingfeng and Pan, Yulin and Liu, Yu},
    journal = {arXiv preprint arXiv:2503.07598},
    year = {2025}
}

@article{cheng2023consistent,
  title={Consistent video-to-video transfer using synthetic dataset},
  author={Cheng, Jiaxin and Xiao, Tianjun and He, Tong},
  journal={arXiv preprint arXiv:2311.00213},
  year={2023}
}

@inproceedings{barron2022mip,
  title={Mip-nerf 360: Unbounded anti-aliased neural radiance fields},
  author={Barron, Jonathan T and Mildenhall, Ben and Verbin, Dor and Srinivasan, Pratul P and Hedman, Peter},
  booktitle={Proceedings of the IEEE/CVF conference on computer vision and pattern recognition},
  pages={5470--5479},
  year={2022}
}

@inproceedings{zhao2025drivedreamer,
  title={Drivedreamer-2: Llm-enhanced world models for diverse driving video generation},
  author={Zhao, Guosheng and Wang, Xiaofeng and Zhu, Zheng and Chen, Xinze and Huang, Guan and Bao, Xiaoyi and Wang, Xingang},
  booktitle={Proceedings of the AAAI Conference on Artificial Intelligence},
  volume={39},
  number={10},
  pages={10412--10420},
  year={2025}
}
}


\end{document}